\documentclass{new_tlp}

\usepackage{times}
\usepackage{helvet}
\usepackage{courier}
\usepackage{graphics}
\usepackage{amssymb}
\usepackage{amsmath}
\usepackage{latexsym}
\usepackage{booktabs}
\usepackage{xcolor}

\usepackage{mdwlist}

\newcommand\bcmdtab{\noindent\bgroup\tabcolsep=0pt%
  \begin{tabular}{@{}p{10pc}@{}p{20pc}@{}}}
\newcommand\ecmdtab{\end{tabular}\egroup}

\newcommand{\ud}{\mathsf{undef}}

\newtheorem{theorem}{Theorem}
\newtheorem{definition}[theorem]{Definition}

\newcommand{\CalP}{{\cal P}}
\newcommand{\CalE}{{\cal E}}
\newcommand{\CalO}{{\cal O}}
\newcommand{\CalA}{{\cal A}}

\newcommand{\CalI}{{\cal I}}
\newcommand{\CalC}{{\cal C}}

\newcommand{\IC}{\CalI \CalC}

\newcommand{\Luka}{\mbox{\tiny\L}} 
\newcommand{\udf}{\mathsf{U}}
\newcommand{\atoms}{\mathsf{atoms}}
\newcommand{\WComp}{\mathsf{wc}\,}

\newcommand{\Lmwcl}{\mathsf{lm}_{\Luka}\WComp}

\newcommand{\ModelsLMWC}{\models_{\Luka}^{lmwc}}

\newcommand{\inspect}{\mathsf{inspect}} 
\newcommand{\insp}{\mathit{insp}}

\newcommand{\add}{\mathit{add}}
\newcommand{\nadd}{\mathit{add^\prime}}
\newcommand{\inex}{\mathit{inex}}
\newcommand{\cig}{\mathit{cig}}
\newcommand{\ab}{\mathit{ab}}

\newcommand{\storm}{\mathit{storm}}
\newcommand{\lightning}{\mathit{lightning}}
\newcommand{\tempest}{\mathit{tempest}}
\newcommand{\ffire}{\mathit{ffire}}
\newcommand{\barbecue}{\mathit{barbecue}}
\newcommand{\dry}{\mathit{dry\_leaves}}
\newcommand{\drys}{\mathit{dry}}
\newcommand{\rained}{\mathit{rained}}
\newcommand{\smoke}{\mathit{smoke}}
\newcommand{\fire}{\mathit{fire}}
\newcommand{\ffighters}{\mathit{ffighters}}
\newcommand{\sirens}{\mathit{sirens}}

  \title[]{Contextual Abductive Reasoning with Side-Effects}
          
  \author[]
         { {\large  \bf Lu{\'i}s Moniz Pereira$^{\mbox{\small \thanks{lmp@fct.unl.pt}} 1}$}   \\ \medskip
         {\large  \bf Emmanuelle-Anna Dietz $^{\mbox{\small \thanks{$\{$dietz,sh$\}$@iccl.tu-dresden.de}} 1, 2}$ } \\ 
         {\large  \bf Steffen H{\"o}lldobler$^{\dag 2}$}  \medskip \\
         $^1$Centro de Intelig\^encia Artificial (CENTRIA),
Departamento de Inform{\'a}tica \\
Faculdade de Ci\^encias e Tecnologia, Universidade Nova de Lisboa, 2829-516 Caparica, Portugal
\medskip\\
$^2$ International Center for Computational Logic \\
TU Dresden, D-01062 Dresden, Germany
}

\begin{document}

\maketitle

\begin{abstract}
The belief bias effect is a phenomenon which occurs when we think that we judge an argument based on our reasoning, 
but are actually influenced by our beliefs and prior knowledge. Evans, Barston and Pollard carried out a psychological 
syllogistic reasoning task to prove this effect. Participants were asked whether they would accept or reject a given 
syllogism. We discuss one specific case which is commonly assumed to be believable but which is actually not
logically valid. By introducing abnormalities, abduction and background knowledge, we adequately model this case under 
the
weak completion semantics. Our formalization reveals new
questions about possible extensions in abductive reasoning. 
For instance, observations and their explanations might include some relevant prior abductive 
contextual information concerning some side-effect
or leading to a contestable or refutable side-effect. A weaker notion indicates the support of some relevant 
consequences by a prior abductive context. 
Yet another
definition describes jointly supported relevant consequences, which captures the idea of two observations containing 
mutually supportive side-effects.
Though motivated with and exemplified by the running psychology application, the various new general abductive context 
definitions are introduced here and given a declarative semantics for the first time, and have a much wider scope of 
application.
Inspection points, a concept introduced by Pereira and Pinto,
allows us to express these definitions syntactically
and intertwine them into an operational semantics.  	 	
\end{abstract}

\section{Introduction}

In the context of abductive reasoning, whenever discovering abductive
explanations for some given primary observation, one may wish to check
too whether some other given additional secondary observations are
true, as logical consequences of the abductive explanations found for
the primary observation. In other words, whether the secondary
observations are plausible in the abductive context of the primary
one, is a common scientific reasoning task.  Thus, for example, one
may attempt to find abductive explanations for such secondary
observations strictly within the context of the given abductive
explanations found for the primary observation -- that is, disallowing
new abductions -- or, nevertheless, allowing additional abductions as
long as they are consistent with the primary ones.
As it were, the explanations of secondary observational consequences may be
consumers, but not producers, of the abductions produced in explaining
the primary observation.
We show this type of reasoning requires the characterization of a new
abduction concept and mechanism, that of \textit{contextual abduction}. We
examine and formalize its variants, and employ these to understand and
justify belief bias of human reasoning in addressing syllogisms.

Our starting point is a psychological study carried out by
\citeN{evans:1983} about 
deductive 
reasoning which demonstrated possibly conflicting processes in 
human reasoning. Participants were presented different syllogisms and had to decide whether they were 
logically valid. 
Consider one of them,~\textsc{S$_{\add}$}: \\ 
 \begin{tabular}{@{\hspace{0mm}}l@{\hspace{1.2mm}}l} 
\textsc{Premise1} & \textit{No addictive things are 
inexpensive. }\\
\textsc{Premise2} &\textit{Some cigarettes are inexpensive.} \\
\textsc{Conclusion} & \textit{Therefore, some addictive things} \textit{are not cigarettes.} 
\end{tabular}
\vspace{0.25mm}
Even though the conclusion does not necessarily follow from the 
premises,\footnote{
In abstract, it could be conceived that all addictive things would be just the (expensive) cigarettes. 
It does not follow because the conclusion supposes an existence which is not warranted by the premises. 
}
participants assumed the syllogism to be logically valid.  
They were explicitly asked
to logically validate or invalidate these syllogisms, but didn't seem to have the intellectual capability to do so. 
Even worse, they were not at all aware about their inabilities. Repeatedly, the majority of people were proffering with 
certainty of confidence the wrong answer. 
\citeNS{evans:1983} concluded that this happened because they were being unduly influenced 
by their own beliefs, their belief bias. 
In~\cite{evans2010thinkingtwice,evans:2012}, the conflict between logic and belief in human
reasoning is discussed extensively.

We will show how the belief bias effect can be explained 
by abductive reasoning and its corresponding side-effects. This forms the basis for
investigating contextual side-effects, (strict) possible side-effects, contextual 
contestable side-effects, and (jointly supported) contextual relevant consequences.

\section{Preliminaries}

We define the necessary notations, simplified for our present
purposes but generalizable as usual (cf. \citeNS{kowalski:2011}).  
We restrict ourselves to datalog programs, i.e.\ the set of terms consists
only of constants and variables.  A (first-order) logic program
$\CalP$ is a finite set of clauses:
\[
A \leftarrow A_1 \wedge \ldots \wedge A_n \wedge \neg{B_1} \wedge \ldots \wedge\neg{B_m},
\]
where $A$ and $A_i$, $1 \leq i \leq n$, are atoms and
$\neg{B_j}$, $1 \leq j \leq m$, are negated atoms. $A$ is
the head and $A_1 \wedge \ldots \wedge A_n \wedge \neg{B_1}
\wedge \ldots \wedge \neg{B_m}$ is the body of the clause.  In
the sequel we will abbreviate the body of a clause by simply writing
$body$.  $A \leftarrow \top$ and $A \leftarrow \bot$ are special cases
of clauses denoting positive and negative facts, respectively.
If an argument is written with an upper case letter, it is a variable;
otherwise it is a constant.
In the sequel, if not denoted otherwise, we assume $\CalP$ to be ground, containing all the ground instances of its clauses.
The set of all atoms occurring in $\CalP$ is $\atoms(\CalP)$.
An atom is undefined in $\CalP$ if it is not the head 
of some clause in $\CalP$ and the corresponding set of these atoms is $\ud(\CalP)$.

\subsection{Three-Valued {\L}ukasiewicz Semantics} \label{sub:3valuedL}

We use the three-valued \citeNS{lukasiewicz:20} semantics.
Table~\ref{tab:3vldluka} defines the corresponding truth tables. 
Interpretations are represented by pairs $\langle I^\top, I^\bot\rangle$, such that
\[
\begin{array}{lll}
I^\top =  \{A \in B_\CalP \mid A \mbox{ is mapped to } \top \} \mbox{   and   }
 I^\bot = \{A \in B_\CalP \mid A \mbox{ is mapped to } \bot \},
\end{array}
 \]
where $\mathcal B_\CalP$ is the Herbrand base with respect to a given program $\CalP$. 
A model of $\CalP$ is an interpretation which maps each clause occurring in $\CalP$ to $\top$. 

One should observe that in contrast to two-valued logic, $A \leftarrow
B$ and
$A \vee \neg B$ are not equivalent under three-valued {\L}ukasiewicz Semantics. 
Consider, for instance, an interpretation~$I$ such that
$I(A) = I(B) = \udf$. Then, $I(A \vee \neg B) = \udf$ whereas
$I(A \leftarrow_{\Luka} B) = \top$.

\subsection{Weak Completion Semantics}

Weak completion semantics has first been introduced by~\citeNS{hk:2009a}
and seems to adequately model~\citeANP{byrne:89}'s \shortcite{byrne:89} suppression 
task~\cite{dietz:hoelldobler:ragni:2012}
and~\citeANP{wason:68}'s~\shortcite{wason:68} selection task~\cite{dietz:hoelldobler:ragni:2013}.
Consider the following transformation for given~$\CalP$: 
\vspace{-0.7cm}
\begin{enumerate*} 
\item[1.] Replace all clauses in $\CalP$ with the
same head $A  \leftarrow body_1, A  \leftarrow
body_2, \dots$ \\ by the single expression~$A  \leftarrow body_1 \lor body_2, \lor \dots$.
\item[2.] if~$A \in \ud(\CalP)$ then add~$A \leftarrow \bot$. 
\item[3.] Replace all occurrences of~$\leftarrow$
by~$\leftrightarrow$. 
\end{enumerate*} \vspace{-0.3cm}
The resulting set of equivalences is called the
\textit{completion} of~$\CalP$~\cite{Clark:1978}.
If Step 2 is omitted, then the
resulting set is called the \textit{weak completion}
of~$\CalP$~($\WComp
\CalP$).

\citeNS{hk:2009b} showed that the model
intersection property holds for weakly completed programs.
This guarantees the existence of a least model for every program.
In computational logic, least models can often be computed as least fixed points of an 
appropriate semantic operator~\cite{apt:emden:1982}.
\citeNS{stenning:vanlambalgen:2008} devised such an operator
which has been generalized for first-order programs by~\citeNS{hk:2009a}: Let~$I$ be an
interpretation and~$\Phi_{\CalP}(I) = \langle J^\top, J^\bot \rangle$,
where \vspace{-0.1cm}
\[\begin{array}{@{\hspace{0mm}}l@{\hspace{0.1mm}}c@{\hspace{0.1mm}}l}    
J^\top & = \{A \mid & \mbox{there exists } A \leftarrow body \in \CalP
\mbox{ with } 
I(body) = \top \},  \\
J^\bot & = \{A  \mid  &\mbox{there exists } A  \leftarrow body \in \CalP
\mbox{
and } \mbox{for all } A \leftarrow body \in\CalP
\mbox{ we
find } I(body) = \bot \}. \\
   \end{array}
\]\vspace{-0.3cm} \\
As shown in \cite{hk:2009a} the least fixed point of~$\Phi_{\CalP}$ 
is identical to the least model of the weak completion
of~$\CalP$~($\Lmwcl\CalP$) under three-valued {\L}ukasiewicz semantics.
Starting with
the empty interpretation~$I = \langle \emptyset, \emptyset
\rangle$,~$\Lmwcl\CalP$ 
can be computed by iterating~$\Phi_{\CalP}$.\footnote{
Weak completion semantics corresponds to well-founded semantics~\cite{gelder:ross:schlipf:1991} for 
tight logic programs~\cite{dietz:hoelldobler:wernhard:2013}.}

\begin{small}
\begin{table*}[t] 
\[
\begin{array}[c]{@{\hspace{0mm}}c|c}
F & \neg {F}\\ \midrule
\top & \bot \\
\bot & \top \\
\udf & \udf \\
\end{array}
\quad\quad
\begin{array}{c|lll}
 \wedge & \top & \udf & \bot \\
\midrule
\top & \top & \udf & \bot \\
\udf & \udf & \udf & \bot \\
\bot & \bot & \bot & \bot \\
 \end{array}
\quad\quad
\begin{array}{c|lll}
 \vee & \top & \udf & \bot \\
\midrule
\top & \top & \top & \top \\
\udf & \top & \udf & \udf \\
\bot & \top & \udf & \bot \\
 \end{array}
\quad\quad
\begin{array}{c|lll} 
\leftarrow_{\Luka} & \top & \udf & \bot \\
\midrule
\top & \top & \top & \top \\
\udf & \udf & \top & \top \\
\bot & \bot & \udf & \top \\
 \end{array}
\quad\quad
\begin{array}{c|ccc}
  \leftrightarrow_{\Luka} & \top & \udf & \bot \\
\midrule
\top & \top & \udf & \bot \\
\udf & \udf & \top & \udf\\
\bot & \bot & \udf & \top \\
 \end{array}
\]
\caption{  $\top$, $\bot$, and $\udf$ denote \textit{true}, \textit{false},
and \textit{unknown}, respectively.\label{tab:3vldluka}}
\end{table*}
\end{small}

\section{Reasoning in an Appropriate Logical Form}

\citeANP{stenning:vanlambalgen:2005}~\shortcite{stenning:vanlambalgen:2005,stenning:vanlambalgen:2008} proposed to 
model 
human 
reasoning by a two step process: Firstly, human reasoning should be modeled 
by setting up an \emph{appropriate representation} and, secondly, the \emph{reasoning 
process} should be modeled
 with respect to this representation. In this section we discuss the first step 
and show how to model syllogisms
 in logic programs.

\subsection{Integrity Constraints} \label{sub:integrity-constraints}

\textsc{Premise1} of \textsc{S$_{\add}$} is 
\begin{center} 
\textit{No addictive things are inexpensive.}  \hfill (1)
\end{center}
and is equivalent to 
\begin{center}
\textit{If something is inexpensive, then it is not addictive.}\hfill (2)
\end{center}
The consequence is the negation of \textit{something is addictive}. 
As weak completion semantics does not allow negative heads in clauses,
for every negative conclusion $\neg p(X)$ we introduce an auxiliary formula
$p^\prime (X)$, which denotes the negation of $p$ and the clause $p (X) \leftarrow \neg{ p^\prime (X)}$. 

We obtain the following preliminary representation of the first premise of~\textsc{S$_{\add}$}  with regard to 
\textit{addictive}: $\nadd (X) \leftarrow \inex(X) $, and $\add(X) \leftarrow \neg{ \nadd (X)}$,  
where $\add(X)$, $\nadd(X)$, and~$\inex(X)$ stand for
$X$ is addictive, not addictive, and inexpensive. 

With the introduction of these auxiliary atoms, the need for integrity constraints arises. 
A least model of the weak completion that contains both $\add(X)$ and $\nadd(X)$ in $I^\top$ 
should be invalidated as a model in general. 
This condition can be represented by a \textit{set of integrity constraints}~$\IC$,
which contains clauses of the following form: $\udf \leftarrow \IC\_Body$, 
where the implication is understood as usual.
For our example above, $\IC_{\add}$ contains one clause: $\udf \leftarrow \add(X) \wedge \nadd(X).$ 
However, the $\Phi$ operator does not consider clauses of this form, thus such denial $\IC$s are not evaluated by it. We 
 apply
a two step approach to take them into consideration. First, we compute the least
model of the given program and second, we verify whether it does satisfy the requirements
of the $\IC$.
Given an interpretation $I$ and a set of integrity constraints $\IC$,
$I$ \textit{satisfies} $\IC$ if and only if all clauses in $\IC$ are true under $I$.
For the following examples, whenever there exists a $p(X)$ and its $p^\prime(X)$ counterpart in $\CalP$,
we implicitly assume that $\IC_p$: $\udf \leftarrow p(X) \wedge p^\prime(X)$.\footnote{This view
on $\IC$s corresponds to the definition
applied for the well-founded semantics in~\cite{Pereira:wfs:iconstraints:91}.}

\subsection{Abnormalities \& Background Knowledge} \label{sub:abnormalities}

A direct representation of \textsc{Premise2} is
\begin{center}
\textit{There exists a cigarette which is inexpensive.}  \hfill (3) 
\end{center}
Additionally, it 
is commonly known that
\begin{center}
\textit{Cigarettes are addictive.} \hfill (4)
\end{center}
%
As discussed in~\cite{evans:1983}, humans seem to have a background knowledge or belief, which, in 
this context, assuming (4), we imply that 
\begin{eqnarray} \label{e:3}
 & \textit{Cigarettes are inexpensive}  \textit{ (compared to other addictive things)} ,   \nonumber 
\end{eqnarray}
which implies~(3) and biases the reasoning towards a representation. The preliminary program representing
the first two premises, is the non-ground program~$\CalP_{add}^{pre}$:
\[ \begin{array}{@{\hspace{0mm}}lll@{\hspace{15mm}}lll}
\nadd (X) & \leftarrow & \inex(X) ,  &  \add (X) & \leftarrow & \neg{ \nadd (X)}, \\
\inex(X) & \leftarrow & \cig(X) , & \cig(a) & \leftarrow & \top. \\
\end{array}
 \]
and its corresponding least model is: \hfill $
 \langle \{ \cig(a), \inex(a), \nadd(a)\}, \{ \add(a) \} \rangle$. \\
This model contradicts the commonly known assumption of~(4).
This background knowledge can be expressed in the program by applying Stenning and van Lambalgen's idea 
to implement conditionals by a normal default permission for implications.This can be
achieved by adding an \textit{abnormality predicate} to the antecedent
of the implication and initially assuming that the abnormality predicate is false.
Following this idea, the initial \textsc{Premise1} in \textsc{S$_{\add}$}
is extended to:  
\begin{center}
\textit{If something is inexpensive and not abnormal, then it is not addictive. 
\\ Nothing (as a rule) is abnormal (regarding \normalfont{(1)}).} 
\end{center}
This 
belief-bias together with the idea to 
represent conditionals by a normal default permission for implication leading to this rendering
\begin{equation*}
\begin{tabular}{@{\hspace{20mm}}c@{\hspace{12mm}}l} 
\textit{If something is a cigarette and not abnormal,} \textit{then it is inexpensive.} & (5) \\
\textit{Nothing (as a rule) is abnormal (regarding \normalfont{(3)})}.
\end{tabular}
\end{equation*}
Together with (4), it leads to 
\begin{center}
 \textit{If something is a cigarette,} \textit{then it is abnormal (regarding \normalfont{(1)}).} 
\end{center}
Finally, the information
in the premises of~\textsc{S$_{\add}$} is encoded as the non-ground program~$\CalP_{\add}$:
\[ \begin{array}{@{\hspace{0mm}}lll@{\hspace{15mm}}lll}
\nadd (X) & \leftarrow & \inex(X) \wedge \neg{ \ab_1 (X)}, & \add (X) & \leftarrow & \neg{ \nadd (X)}, \\
\inex(X) & \leftarrow & \cig(X) \wedge \neg{\ab_2(X)}, &  \ab_2(X) & \leftarrow & \bot, \\
\ab_1(X) & \leftarrow & \bot, & \ab_1(X) & \leftarrow & \cig(X), \\
\cig(a) & \leftarrow & \top. \\
\end{array}
 \]
$\CalP_{\add}$ represents the contextual background knowledge of syllogism S$_{\add}$.

\section{Abduction and Predictability}

In scientific methodology, a prediction can be made~by adding hypotheses to knowledge known about the world. As 
specified by the classical hypothetic-deductive meth\-od~\cite{Hempel:1966}, scientific inquiry is carried out in three 
stages: hypotheses generation, prediction, and evaluation. One or more hypotheses may be generated, by abduction, to 
explain observed events. A generated set of hypotheses (or assumptions) can thence be employed for predicting unseen 
events by means of deduction, on the implicit condition of not making further abductions 
(cf.~\citeNS{pereira:pinto:2011}). 
The predicted events can then hopefully be tested against reality, in the form of such observable deduced 
side-effects, in order to evaluate the plausibility of the set of hypotheses.

Abduction, or inference to the best explanation (its usual designation in the philosophy of science), is a reasoning 
method whereby one chooses those hypotheses that would, if true, best explain observed evidence or enable to satisfy 
some query, whilst meeting attending constraints. 
Abduction has been well studied in the field of computational logic -- and logic programming in particular -- for a few 
decades now. Abduction, when added to logic programs, offers a formalism to declaratively express problems in a variety 
of areas and empowers many applications, e.g.\ in decision-making, diagnosis, planning, belief revision, and 
conditional 
reasoning.
In logic programs, abductive hypotheses (or abducibles) are named given atoms of the program which 
have no rules, and whose truth value is not initially assumed, and hence unknown.

The approach presented in~\cite{torasso:console:luigi:1995} treats 
abduction with completion semantics 
and does not study side-effects. In the following 
we consider abduction with weak completion semantics and introduce the examination of side-effects 
in contexts afforded by abduction.

\subsection{Abductive Framework} \label{sub:abduct-frame}

Following \citeNS{kakas:etal:93} we
consider an \textit{abductive framework} consisting of a
program~$\CalP$ as
knowledge base, a collection of atoms~$\CalA$
of abducibles syntactically represented by the set of the (positive and negative) facts for each
undefined ground
atom in~$\CalP$, a set of integrity constraints $\IC$, and the logical consequence relation~$\ModelsLMWC$,
where~$\CalP \ModelsLMWC F$ if
and only if~$\Lmwcl \CalP (F) = \top$ for the formula~$F$. An
\textit{observation} is a set of (at least one) literals.

The truth value of abducibles may be independently assumed true or false, via either their positive or negated form, as 
the case may be, in order to produce an abductive explanation for an observation -- or solution to a query --, which is 
a consistent set of assumed hypotheses in the form of abducibles. 

An abductive solution is a consistent set of 
abducible instances that, when substituted by their assigned truth value in $\CalP$, affords 
us with a model of $\CalP$ (for the specific semantics used on $\CalP$), which satisfies the observation 
(or query) and any imposed integrity constraints -- a so-called abductive model.
In our notation this amounts to adding to $\CalP$ the corresponding positive and negative facts representing a 
solution's abducibles.

What we dub \textit{observation} is analogous to a \textit{query} whose explanation is desired, not necessarily 
something actually observed.
\begin{definition} \label{abduction}
Let~\normalfont{$\langle \CalP, \CalA, \IC, \ModelsLMWC \rangle$}  be an abductive
framework, $\CalO$ be an observation, and $\CalE$ be an explanation which is a (consistent) subset of $\CalA$,
a set of integrity constraints $\IC$,
  and the consequence relation $\ModelsLMWC$, defined for all formulas~$F$.
 \begin{description}
 \item[$\CalO$] is \textbf{explained by $\CalE$ given $\CalP$ and $\IC$} iff 
 $\CalP \cup \CalE \ModelsLMWC \CalO$, where $\CalP \not \ModelsLMWC \CalO$ and
 $\Lmwcl (\CalP \cup \CalE)$ \textit{satisfies} $\IC$.
\item[$\CalO$] is \textbf{explained given $\CalP$ and $\IC$} iff there exists an
  $\CalE$ such that $\CalO$ is explained by $\CalE$ given $\CalP$ and~$\IC$.
 \end{description} 
In abduction, as for its deduction counterpart, \textit{credulous} and \textit{skeptical reasoning} varieties 
are 
distinguished. Credulous reasoning consists in finding if there exists at least one model of the 
program -- according to some pre-established semantics -- which entails the observation to be explained. 
Skeptical reasoning demands 
that every model of the program entails the observation.
\begin{description}
  \item[$F$] \textbf{follows skeptically from~$\CalP$,~$\IC$ and~$\CalO$} iff
$\CalO$ can be explained given $\CalP$ and $\IC$, and for all
minimal (or some alternative preference criterion instead) explanations~$\CalE$ for $\CalO$ it holds that $\CalP \cup \CalE 
\ModelsLMWC F$.
\item[$F$] \textbf{follows credulously from~$\CalP$, ~$\IC$ and~$\CalO$} iff
there exists a minimal (or some alternative preference criterion instead) explanation~$\CalE$ for $\CalO$ and it holds that 
$\CalP \cup \CalE \ModelsLMWC F$.
\end{description}
\end{definition} 
Because the satisfaction of integrity constraints ($\IC$s) can require abductions, 
we must allow $\IC$s to be actively productive of abductions, and not just use $\IC$s to 
subsequently disallow abductive solutions that invalidate the least model with respect to them. 
However, we cannot actively promote that an atom is required to be false or else unknown, one of the two, in order for 
$\IC$s to be satisfied. 
The only possibility is to
impose its falsity: because
$\IC$s have the form of denials, we can think of $\udf \leftarrow A$ 
as just an atom that we wish must never be explained by any explanation. 

As our underlying semantics is three-valued, 
we may but might not make each undefined atom an abducible, so that we allow for unknown, non-abducible 
susceptible
knowledge, which can be guaranteed by adding the clause~$A \leftarrow A$ for
the predicate under consideration. Furthermore, when making use of skeptical reasoning,
we conclude that an abducible is unknown if it does not have the same binary truth value in all 
models. For instance consider the program $\CalP = \{A \leftarrow B, A \leftarrow C\}$ and observation $\CalO_A = \{A 
\}$,
for which there are two minimal explanation $\CalE_B = \{ B \leftarrow \top \}$ and $\CalE_C = \{ C \leftarrow \top \}$. 
Under skeptical
reasoning $A$ does not follow from all minimal explanations and thus $B$ and $C$ stay unknown.

In previous approaches, weak completion semantics was used for 
cases expressed in propositional logic, and
abduction under skeptical reasoning 
seemed adequate~\cite{HPW:2011:AMHR}. 

\subsection{Usual Contextual Abduction} \label{sub:usual}

One important extension of abduction pertains to the issue that, whenever discovering abductive solutions, i.e.\ 
explanations, for some given primary observation, one may wish to check too whether some other given additional 
secondary observations are true, being a logical consequence of the abductive explanations found for the primary 
observation. In other words, whether the secondary observations are plausible in the abductive context of the primary 
one.
Indeed, often, besides needing to abductively discover which hypotheses to assume in order to satisfy some 
condition, we may also want to know some of the side-effects of those assumptions.

We address the issue of relaxing or loosening the implicit condition about additional abductions not being 
permitted whilst considering the observable side-effects explained by deductive prediction. In other words, prediction 
may be allowed recourse to additional assumptions, but nevertheless must make use of at least some of the initial 
explanations. If several such explanations exist for the observations concerned, then we might
want to define alternative conditions that are less strict with regard to each set or to the collection of 
sets. 
Reuse of contextual abductions, by resorting to an implementation of tabled abduction for well-founded
semantics, is reported in~\cite{saptawijaya:pereira:2013}.

Let's consider again $\CalP_{\add}$, its weak completion consists of the following equivalences:
 \[ \begin{array}{@{\hspace{0mm}}lll@{\hspace{15mm}}lll}
 \add (a) & \leftrightarrow & \neg{ \nadd (a)}, & \nadd (a) & \leftrightarrow & \inex(a) \wedge \neg{ \ab_1(a)}, \\
 \inex(a) & \leftrightarrow & \cig(a) \wedge \neg{\ab_2(a)}, & \ab_2(a) & \leftrightarrow & \bot, \\
  \ab_1(a) & \leftrightarrow & \cig(a),\footnotemark &  \cig(a) & \leftrightarrow & \top . \\
\end{array}
\]\footnotetext{$\bot \vee \cig(a)$ is semantically equivalent to $\cig(a)$
under {\L}ukasiewicz Semantics.}
Its least model, $\Lmwcl\CalP_{\add}$, is $
 \langle \{ \cig(a), \inex(a), \add(a),  \ab_1(a) \},  \{\nadd (a), \ab_2(a) \}\rangle$, 
from which we cannot derive the \textsc{Conclusion} in \textsc{S$_{\add}$}. Obviously,
the \textsc{Conclusion} is  something about an object which is not $a$.
The first part of this conclusion is an observation, let's say about $b$: $\CalO_{\add(b)} = \{\add(b)\}$,
which we need to explain as described in the previous subsection. 
The set of abducibles with respect to $b$ is: 
$\CalA_{ \CalP_{\add} } = \{\cig(b) \leftarrow \top, \cig(b) \leftarrow \bot \}$.
$\CalO_{\add(b)}$ is true if $\nadd (b)$ is false which 
is false if $\inex(b)$ is false or $\ab_1(b)$ is true. Either 
$\inex(b)$ is false but then $\cig(b)$ is false or $\ab_1(b)$ is true but then $\cig(b)$ is true. 
For $\CalO_{\add(b)}$ we have two minimal explanations $\CalE_{\neg{ \cig(b)}} =  \{ \cig(b) \leftarrow \bot\}$ 
and  $\CalE_{\cig(b)} = \{\cig(b) \leftarrow \top\}$. The corresponding least models of the weak completion are:
\[
 \begin{array}{@{\hspace{0mm}}l@{\hspace{1mm}}l@{\hspace{1mm}}l}
  \displaystyle 
  \Lmwcl (\CalP_{\add} \cup \CalE_{\neg{\cig(b)}})&  = &  \langle \{  \dots, 
\add(b), \dots\},  \{  \dots, \cig(b), \inex(b), \dots \} 
\rangle , \\ \displaystyle
  \Lmwcl (\CalP_{\add} \cup \CalE_{\cig(b)}) & = & \langle \{ \dots,
\add(b), \cig(b), \inex(b),\dots\}, \{ \dots \}
\rangle . 
 \end{array}
\]
Under credulous reasoning we conclude, given explanation~$\CalE_{\neg {\cig(b)}}$,
that the \textsc{Conclusion} of \textsc{S$_{\add}$} is true, as there exists
something addictive which is not a cigarette.
I.e., there is a least model which abductively explains that an observed addictive $b$ is not a cigarette.
\\ 
However, we could also explain $\CalO_{\add(b)}$ by $\CalE_{\cig(b)}$, which is an equally justified explanation.
We would prefer our formalism to reflect that the first premise describes the usual and
the second premise describes the exceptional case. That is, an inexpensive cigarette is meant to be the exception not the rule in
the context of things that are addictive. This exceptional case should then only
be considered when more is known about $b$. 
This preference is not expressed in our framework yet. 
The following section discusses and proposes a solution to this issue.

\section{Inspection Points} \label{sect:inspection}

Until now, we assumed the possible observations as given, and 
easily identified from the corresponding context which was the exceptional and which the usual case. 
However, this is not explicitly encoded in our 
logic programs yet, and needs be syntactically indicated.
For this purpose we investigate
and apply inspection points, originally presented by~\citeNS{pereira:pinto:2011}.

In~\cite{pereira:pinto:2011}, the authors present the concept of inspection points in abductive logic programming
and show how one can employ it to investigate side-effects of interest in order to help
choose among abductive solutions. In what follows we discuss this approach and show how inspection points can be 
modeled for our previous examples accordingly, in what concerns abducibles.
Given an atom $A$, we introduce the following two reserved (meta-)predicates:
\[
 \inspect(A)  \hspace{1cm} \mbox{ and } \hspace{1cm} \inspect_\neg(A),
\]
which are special cases of abducibles.
 They differ from the usual abducibles
in the way that they can only be abduced whenever $A$ or $\neg A$ have been abduced somewhere 
else already. That is, an abductive solution or explanation $\CalE$ 
is only valid when for each $\inspect(A)$ (respectively $\inspect_\neg(A)$) it contains,
it also contains a corresponding $A$ (respectively $\neg A$). That is, in a solution the consumers,
here represented by the inspection points $\inspect(A)$ and $\inspect_\neg(A)$, respectively, 
must have a matching producer. The producers correspond to the usual abducibles.

One should observe that for a treatment of inspection points for all literals in a program and not just the abducible 
ones, we would simply need to adopt the program transformation technique of~\citeNS{pereira:pinto:2011}, which 
recursively relays inspection of non-terminal literals to the base inspection of terminals.
\\
Let us consider again the example of Section~\ref{sub:usual} where we stated that we would
prefer to distinguish between the usual case and the exceptional case. 
We can
now represent this in our logic program 
by replacing our $\ab_{1}$-clause accordingly. The new non-ground program, $\CalP_{\add}^{\insp}$, is:
\[
  \begin{array}{l@{\hspace{1mm}}l@{\hspace{1mm}}l}
 (\CalP_{\add}  \setminus
   \{ \ab_{1}(X) \leftarrow \cig(X) \})   \cup \    \{ \ab_{1}(X) \leftarrow \inspect(\cig(X)) \}.
  \end{array}
\]
Suppose, $b$ is addictive, i.e.\ $\CalO_{\add(b)}$. As $\cig(b)$ is undefined, $\inspect(\cig(b))$ becomes false,
and~$\ab_{1}(b)$ will be false rather than unknown, that is, its falsity is obtained because nothing is known about 
$\cig(b)$. The only minimal explanation for $\CalO_{\add(b)}$ is now generated by~$\inex(b)$ being false, which
is achieved by~$\Lmwcl (\CalP_{\add}^{\insp} \cup \CalE_{\neg { \cig(b)}})$: 
\[
\langle \{ \dots,
\add(b), \dots\},  \{ \dots, \nadd (b), \inspect(\cig(b)),   \ab_1(b), \inex(b), \ab_2(b), \dots\} \rangle.
\]
Even under skeptical reasoning, there exists an addictive thing which is not a cigarette.

\section{The Expressive Power of Inspection Points in Contextual Abduction}

Inspection points allow us to specify various definitions, which we will provide
-- relative to explanation $\CalE$ of observation $\CalO$, set of 
abducibles $\CalA$ and background knowledge $\CalP$ --, to the effect of relaxing the impermissibility of convoking 
additional hypotheses to explain side-effects in context $\CalE$. 
Consider the following program,
$\CalP_{\fire}$:
\[\begin{array}{@{\hspace{0mm}}lll@{\hspace{10mm}}lll@{\hspace{0mm}}}
\storm & \leftarrow & \lightning \wedge \neg \ab_1, & \ffire & \leftarrow & \inspect(\lightning) \wedge \neg \ab_3,\\
     \storm & \leftarrow & \tempest \wedge \neg \ab_2, & \ffire  & \leftarrow & \barbecue \wedge \neg \ab_3,  \\
     \ab_3 & \leftarrow & \neg \dry,  & \rained & \leftarrow & \inspect_\neg(\dry) \wedge \neg \ab_4, \medskip \\
    \smoke & \leftarrow & \fire \wedge \inspect(\ffighters), & \sirens & \leftarrow & \inspect(\fire) \wedge \ffighters,\medskip  \\ 
    \ab_1 & \leftarrow &  \bot,\hfill \ab_2 \hspace*{0.3cm} \leftarrow \hspace*{0.3cm}  \bot,   & \ab_3 & \leftarrow & \bot, 
    \hfill \ab_4 \hspace*{0.3cm} \leftarrow \hspace*{0.3cm}  \bot.  \\ 
     \end{array}
\] 
where $\ffire$ means forest fire and $\ffighters$ means fire fighters.
The set of abducibles, $\CalA_{\CalP_{\fire}}$, is:
\[
 \begin{array}{@{\hspace{0mm}}lll@{\hspace{15mm}}lll@{\hspace{0mm}}}
  \tempest & \leftarrow & \top, & \tempest & \leftarrow & \bot, \\
  \barbecue & \leftarrow & \top, & \barbecue & \leftarrow & \bot,  \medskip \\
  \lightning & \leftarrow & \top, & \lightning & \leftarrow & \bot, \\
  \inspect(\lightning) & \leftarrow & \top, & \inspect(\lightning) & \leftarrow & \bot, \medskip \\
  \dry & \leftarrow & \top, & \dry & \leftarrow & \bot, \\
  \inspect_\neg(\dry) & \leftarrow & \top, & \inspect_\neg(\dry) & \leftarrow & \bot, \medskip \\
  \fire & \leftarrow & \top, & \fire & \leftarrow & \bot, \\
  \inspect(\fire) & \leftarrow & \top, & \inspect(\fire) & \leftarrow & \bot,\medskip \\
  \ffighters & \leftarrow & \top, & \ffighters & \leftarrow & \bot,  \\
  \inspect(\ffighters ) & \leftarrow & \top, & \inspect(\ffighters ) & \leftarrow & \bot. \\
 \end{array}
\]
For simplicity, we assume that $\IC_{\fire} = \emptyset$.
In the sequel, we provide definitions and
 abductive examples which clarify how inspection points enrich the expressiveness
 of logic programs.

\subsection{Contextual Side-effects} \label{sub:sideeffects}

Consider the following definition which includes various notions of contextual side-effects:
\begin{definition}\label{sideffects-inspect-n}
Given background knowledge $\CalP$ and set of integrity constraints $\IC$, let $\CalO_1$ be an observation, 
$\CalE_1$ be an explanation of $\CalO_1$, $\CalO_2$ be an observation, and $\CalE_2$ be an explanation of $\CalO_2$.
\begin{description}
 \item[$\CalO_2$] is a \textbf{necessary contextual side-effect of
     $\CalO_1$ given~$\CalP$ and $\IC$} iff for all $\CalE_1$ there exists $\CalE_2$ such that, for any
     $\inspect(A)$, $\inspect_\neg(A)$~$\in~\CalE_2$ for which respectively
     $A$, $\neg A \not \in \CalE_2$, and some such exists in $\CalE_2$, then, $A$, $\neg A \in \CalE_1$,
     accordingly.
 \item[$\CalO_2$] is a \textbf{strict necessary contextual side-effect of $\CalO_1$ given~$\CalP$ and $\IC$} iff 
 $\CalO_2$ is a necessary contextual side-effect of $\CalO_1$ given $\CalP$, and $\CalE_1 \subseteq \CalE_2$ for any 
$\CalE_1$.
\item[$\CalO2$] is a \textbf{possible contextual side-effect of
    explained $\CalO_1$ given~$\CalP$ and $\IC$} iff there exists an $\CalE_1$ and an $\CalE_2$ such that for any
     $\inspect(A)$, $\inspect_\neg(A)$~$\in~\CalE_2$ for which respectively
     $A$, $\neg A \not \in \CalE_2$, and some such exists in $\CalE_2$, then $A$, $\neg A \in \CalE_1$, accordingly.
\item[$\CalO_2$] is a \textbf{strict possible contextual side-effect of
$\CalO_1$ given~$\CalP$ and $\IC$}  iff 
 $\CalO_2$ is a necessary contextual side-effect of $\CalO_1$ given $\CalP$, and for some $\CalE_1$, $\CalE_1 \subseteq 
\CalE_2$.
\end{description}
\end{definition}
The idea behind the definition for necessary contextual side-effects is
that every explanation $\CalE_1$ for $\CalO_1$ affords us with one \textit{complete} explanation by which 
any (inspection) \textit{incomplete} explanation $\CalE_2$ for $\CalO_2$ can be necessarily completed.
The idea behind the definition for possible contextual side-effects is
that there is some explanation $\CalE_1$ for $\CalO_1$ which affords us with one \textit{complete} explanation 
by which some (inspection) \textit{incomplete} explanation $\CalE_2$ for $\CalO_2$ can be completed.

Consider again $\CalP_{\fire}$, where there are two rules
for $\ffire$: one explanation is $\barbecue$ and another one is
$\lightning$. However, we assume that a lightning causing a forest fire is much more unlikely than a barbecue,
and therefore, $\lightning$, when not observed directly, only counts
as a plausible explanation when it has been abduced by some other observation. This is expressed 
by $\inspect(lightning)$. 
Assume that we observe a storm and know that nothing is abnormal with respect to $\ab_3$, thus
the leaves are dry: $\CalO_{\storm, \drys} = \{ \storm, \dry \}$. 
The minimal explanations
are $\CalE_{\lightning} = \{ \lightning \leftarrow \top \} $ and $\CalE_{\tempest} = \{ \tempest 
\leftarrow \top\}$.
We have only enough evidence to
explain $\CalO_{\ffire}$ if we abduce $\CalE_{\lightning}$ for $\CalO_{\storm, \drys}$.
By Definition~\ref{sideffects-inspect-n}, $\CalO_{\ffire}$ is a 
strict possible contextual side-effect
of explained $\CalO_{\storm, \drys}$ given~$\CalP_{\fire}$ and $\IC_{\fire}$.

\subsection{Contestable Contextual Side-effects} \label{sub:contestable}

Whereas up till now we stipulated cases where known explanations gave grounds for strengthening the 
plausibility of contextual side-effects, we next turn our attention to cases where the latter are to some extent 
contested and made implausible, by appealing to their negation.
Analogously to Definition~\ref{sideffects-inspect-n}, its counterpart, contestable contextual side-effects, is defined as follows:
\begin{definition}\label{contested-sideffects-inspect-n}
Given background knowledge $\CalP$ and set of integrity constraints $\IC$, let $\CalO_1$ be an observation, 
$\CalE_1$ be an explanation of $\CalO_1$, $\CalO_2$ be an observation, and $\CalE_2$ be an explanation of $\neg \CalO_2$.
\begin{description}
 \item[$\CalO_2$] is a \textbf{necessarily contested contextual
     side-effect of $\CalO_1$ given $\CalP$ and $\IC$} iff 
 for all explanations $\CalE_1$ there exists $\CalE_2$ such that, 
 for any $\inspect(A)$, $\inspect_\neg(A)$~$\in~\CalE_2$, for which respectively
     $A$, $\neg A \not \in \CalE_2$, and some such exists in $\CalE_2$, then, $A$, $\neg A \in \CalE_1$,
     accordingly.  
 \item[$\CalO_2$] is a \textbf{possibly contested contextual side-effect of $\CalO_1$ given $\CalP$ and $\IC$} iff
 there exists an $\CalE_1$ and an $\CalE_2$ such that for any
     $\inspect(A)$, $\inspect_\neg(A)$~$\in~\CalE_2$ for which respectively
     $A$, $\neg A \not \in \CalE_2$, and some such exists in $\CalE_2$, then $A$, $\neg A \in \CalE_1$, accordingly.
\end{description}
\end{definition}
The idea behind necessarily contested contextual side-effects, is, 
that every explanation $\CalE_1$ for $\CalO_1$ affords us with one \textit{complete} explanation under which some
\textit{incomplete} explanation $\CalE_2$ for $\neg{\CalO_2}$ is necessarily completed.
The idea behind the definition for the possibly contested contextual side-effects, is,
that there is at least one explanation $\CalE_1$ for $\CalO_1$ which affords us 
with one \textit{complete} explanation under which some \textit{incomplete} 
explanation $\CalE_2$ for $\neg{\CalO_2}$ is necessarily completed.

Consider $\rained \leftarrow \inspect_\neg(\dry)$ in $\CalP_{\fire}$: it states that 
if, for some other observation we explained that the leaves are not dry, then it
rained. Thus, when we observe a forest fire, 
then, one part of the abduced explanation will be, in any case, independent of whether
there was a lightning or a barbecue, that the leaves are dry. 
However this explanation will lead to $\inspect_\neg(\dry)$ being false, which makes $\rained$ false as well.
$\CalO_{\neg {\rained}}$ is a consequence that follows from the explanation for
$\CalO_{\ffire}$.
According to Definition~\ref{contested-sideffects-inspect-n}, $\CalO_{\neg{\rained}}$ is a
necessarily contested contextual side-effect of $\CalO_{ff}$ given $\CalP_{\fire}$ and $\IC_{\fire}$.

Another variation of contestable side-effects is abductive rebuttal. In this case, the side-effect
directly contradicts an \text{observation}. That is to say, it is the case that $\CalO_2 \equiv \neg{
\CalO_1}$ in Definition~\ref{contested-sideffects-inspect-n}. 
The second observation, $\CalO_2$ has to be the head of some clause and the negation of $\CalO_1$.\footnote{As
negative heads in our programs are not allowed, we can model these cases as
described in Section~\ref{sub:integrity-constraints}.}

\subsection{Contextual Relevant Consequences}

We identify two notions of contextual relevant consequences and define them as follows:
\begin{definition}\label{consequence-inspect}
Given background knowledge $\CalP$ and set of integrity constraints $\IC$, let $\CalO_1$ be an observation, 
$\CalE_1$ be an explanation of $\CalO_1$, $\CalO_2$ be an observation, and $\CalE_2$ be an explanation of $\CalO_2$.
 \begin{description}
 \item[$\CalO_2$] is a \textbf{necessary contextual relevant consequence of $\CalO_1$ 
given $\CalP$ and $\IC$}
iff for all $\CalE_2$ there exists $\CalE_1$ consistent with 
 $\CalE_2$ such that there exists either $\inspect(A)$ or $\inspect_\neg(A)$~$\in~\CalE_2$ for which respectively
     $A$, $\neg A \not \in \CalE_2$, but in $\CalE_1$.
   \item[$\CalO_2$] is a \textbf{possible contextual relevant consequence of $\CalO_1$ given $\CalP$ and $\IC$}
 iff there exists $\CalE_2$ there exists $\CalE_1$ consistent with 
 $\CalE_2$ such that there exists either $\inspect(A)$ or $\inspect_\neg(A)$~$\in~\CalE_2$ for which respectively
     $A$, $\neg A \not \in \CalE_2$, but in $\CalE_1$.
\end{description}
\end{definition}
Assume we only observe $\CalO_{\storm}$ given $\CalP_{\fire}$ and do not know whether the 
leaves are dry.  Then $\CalE_{\lightning} = \{ \lightning \leftarrow \top\}$ is
only partly explaining $\CalO_{\ffire}$, as additionally we need to abduce
that the leaves are dry. However, as this explanation 
is not decorated with the inspect predicate, it can be abduced straightforwardly by $\CalO_{\ffire}$.
Accordingly, 
$\CalO_{\ffire}$ is a possible contextual 
relevant consequence of $\CalO_{\storm}$ given $\CalP^{\fire}$ and $\IC_{\fire}$.
The case of contested relevant contextual consequence, 
where mere intersection is the case, could too be analogously expressed. 

\subsection{Jointly Supported Contextual Relevant Consequences} \label{sect:jointly}


It might be the case that two observations contain side-effects of each other, simultaneously. 
That is, more generally, we can allow for each observation to need inspection of the abducibles of the other 
observation; that is, they are mutually plausibly explained together, but not each by itself!
\begin{definition} \label{jointlysupported-inspect}
Given background knowledge $\CalP$ and set of integrity constraints $\IC$, let $\CalO_1$, $\CalO_2$ be observations.
\begin{description}
 \item[$\CalO_1$] and $\CalO_2$ are \textbf{necessarily jointly
     supported contextual relevant consequences given $\CalP$ and $\IC$} iff 
 $\CalO_1$ is a necessary contextual relevant consequence of $\CalO_2$ given $\CalP$ and $\IC$, and
 $\CalO_2$ is a necessary contextual relevant consequence of $\CalO_1$ given~$\CalP$ and $\IC$ as defined in 
Definition~\ref{consequence-inspect}.
 \item[$\CalO_1$] and $\CalO_2$ are \textbf{possibly jointly supported
     contextual relevant consequences given $\CalP$ and $\IC$} 
iff 
 $\CalO_1$ is a possible contextual relevant consequence of $\CalO_2$ given $\CalP$ and $\IC$, and
 $\CalO_2$ is a possible contextual relevant consequence of $\CalO_1$ given $\CalP$ and $\IC$, as defined in 
Definition~\ref{consequence-inspect}.
\end{description}
\end{definition}
Let us observe $\CalO_{\smoke}$ in $\CalP_{fire}$. Then we can abduce $\fire$,
but $\ffighters$ needs to be explained by some other observation. On the other hand,
by observing $\CalO_{\sirens}$, we can abduce $\ffighters$ but not $\fire$. 
Accordingly, $\CalO_{smoke}$ and $\CalO_{sir}$ are necessarily jointly
     supported contextual relevant consequences given $\CalP^{fire}$ and $\IC_{\fire}$.

\section{Conclusion and Future Work}
Weak completion semantics is based on a previously proposed approach that seems to adequately model
the Wason selection task and Byrne's suppression task. 
Yet it seems to adequately model another
human reasoning task which includes the belief-bias effect.
Taking our formalization as starting point, we showed by running examples the need and use for 
possible extensions in abductive reasoning. Introducing the concept of inspection points in our current framework
by applying reserved (meta-)predication for all abductive ground atoms,
makes it possible to differentiate between consumed and produced abducibles. This distinction 
allows us to 
implement the concepts of contextual side-effect, (jointly supported) contextual relevant consequences and contestable 
contextual side-effects.
Belief bias can thus be modeled using mechanisms described and formalized more abstractly, 
which deal with contextual abductive reasoning by means of taking side-effects under consideration, 
applicable to a larger scope of problems, via the notion of inspection points (references to other published examples
 are given).
Our abstract formalism opens therefore the way to a wider use, not restricted to psychological modeling.
 \\
When examining abductive explanatory plausibility and contextual counterfactual 
reasoning new
questions raise on whether new observations should be explained by possible or by necessary side-effects. They 
again might explain
new and possibly unexpected side-effects. 
Additionally, we need to explore how to deal with inconsistency.
Another aspect is about choosing the most appropriate explanations.
In our examples, we follow Occam's razor with respect to competing explanations, that is, we only consider the minimal 
explanations. 
However, there are other measures of preferences which might be more appropriate (e.g. from a human reasoning point of 
view) and which should be further investigated.
\\
In a future extension, some abducibles can be abduced in a three-valued way if we're trying to make the top goal 
unknown. This is typical of fault-finding, which concerns two separate problems: (1) finding the abducibles which, if 
unknown for some faulty components, would make unknown the predicted output using the correct model of the 
artifact, that happens to be at odds with the faulty output behavior of the artifact; (2) abducing the faulty behavior of 
components (using a model of the artifact comprising the modeling of faults) to actually predict the faulty behavior. 
This requires abducing normality of a component as a default, and 
going back over that normality to change it to unknown, whose technicalities are beyond the scope of this paper.
\\
In a psychological modeling case, the artifact we might be aiming to find faults about can be some 
psychological model. We then may have available some real person's specific behavior that the model is not consistent 
with, i.e. it wrongly predicts the negation of that behavior. Consequently, we want at least render our model 
consistent with that behavior by forcing some abducible or other to be unknown, and thus its specific prediction is 
unknown, rather than the 
negation of the person's behavior. A second step is to improve and correct the model to make the right prediction. This 
is comparable to introducing a component's faulty behavior model into a correct model of the artifact, whereas now the 
faulty artifact is the (incorrect) model, and the missing faulty component model refers to a missing model part that 
would produce the correct behavior prediction.

\section{Acknowledgments}
We thank Pierangelo Dell'Acqua for his comments. This work was partly funded by DAAD's IPID program, 
financed by the German Federal Ministry of Education and Research~(BMBF).

\bibliographystyle{acmtrans}
\bibliography{bib}

\begin{thebibliography}{}

\bibitem[\protect\citeauthoryear{Apt and van Emden}{Apt and van
  Emden}{1982}]{apt:emden:1982}
{\sc Apt, K.~R.} {\sc and} {\sc van Emden, M.~H.} 1982.
\newblock Contributions to the theory of logic programming.
\newblock {\em Journal of the ACM\/}~{\em 29,\/}~3, 841--862.

\bibitem[\protect\citeauthoryear{Byrne}{Byrne}{1989}]{byrne:89}
{\sc Byrne, R. M.~J.} 1989.
\newblock {Suppressing valid inferences with conditionals}.
\newblock {\em Cognition\/}~{\em 31}, 61--83.

\bibitem[\protect\citeauthoryear{Clark}{Clark}{1978}]{Clark:1978}
{\sc Clark, K.~L.} 1978.
\newblock Negation as failure.
\newblock In {\em Logic and Data Bases}, {H.~Gallaire} {and} {J.~Minker}, Eds.
  Vol.~1. Plenum Press, New York, NY, 293--322.

\bibitem[\protect\citeauthoryear{Dietz, H\"{o}lldobler, and Ragni}{Dietz
  et~al\mbox{.}}{2012}]{dietz:hoelldobler:ragni:2012}
{\sc Dietz, E.-A.}, {\sc H\"{o}lldobler, S.}, {\sc and} {\sc Ragni, M.} 2012.
\newblock A computational logic approach to the suppression task.
\newblock In {\em Proceedings of the 34th Annual Conference of the {C}ognitive
  {S}cience {S}ociety, {CogSci~2013}}, {N.~Miyake}, {D.~Peebles}, {and} {R.~P.
  Cooper}, Eds. {Austin, TX: Cognitive Science Society}, 1500--1505.

\bibitem[\protect\citeauthoryear{Dietz, H\"{o}lldobler, and Ragni}{Dietz
  et~al\mbox{.}}{2013}]{dietz:hoelldobler:ragni:2013}
{\sc Dietz, E.-A.}, {\sc H\"{o}lldobler, S.}, {\sc and} {\sc Ragni, M.} 2013.
\newblock A computational logic approach to the abstract and the social case of
  the selection task.
\newblock In {\em Proceedings of the 11th International Symposium on Logical
  Formalizations of Commonsense Reasoning, {COMMONSENSE~2013}}. Aeya Nappa,
  Cyprus.

\bibitem[\protect\citeauthoryear{Dietz, H\"{o}lldobler, and Wernhard}{Dietz
  et~al\mbox{.}}{2013}]{dietz:hoelldobler:wernhard:2013}
{\sc Dietz, E.-A.}, {\sc H\"{o}lldobler, S.}, {\sc and} {\sc Wernhard, C.}
  2013.
\newblock Modeling the suppression task under weak completion and well-founded
  semantics.
\newblock {\em Journal of Applied Non-Classical Logics\/}.

\bibitem[\protect\citeauthoryear{Evans}{Evans}{2010}]{evans2010thinkingtwice}
{\sc Evans, J.} 2010.
\newblock Reasoning and imagination.
\newblock In {\em Thinking Twice: Two Minds in One Brain}. OUP Oxford.

\bibitem[\protect\citeauthoryear{Evans}{Evans}{2012}]{evans:2012}
{\sc Evans, J.} 2012.
\newblock Biases in deductive reasoning.
\newblock In {\em Cognitive Illusions: A Handbook on Fallacies and Biases in
  Thinking, Judgement and Memory}, {R.~Pohl}, Ed. Psychology Press.

\bibitem[\protect\citeauthoryear{Evans, Barston, and Pollard}{Evans
  et~al\mbox{.}}{1983}]{evans:1983}
{\sc Evans, J.}, {\sc Barston, J.~L.}, {\sc and} {\sc Pollard, P.} 1983.
\newblock On the conflict between logic and belief in syllogistic reasoning.
\newblock {\em Memory \& Cognition\/}~{\em 11,\/}~3, 295--306.

\bibitem[\protect\citeauthoryear{Hempel}{Hempel}{1966}]{Hempel:1966}
{\sc Hempel, C.~G.} 1966.
\newblock {\em Philosophy of Natural Science}.
\newblock Prentice Hall, Englewood Cliffs, NJ.

\bibitem[\protect\citeauthoryear{H{\"o}lldobler and {Kencana
  Ramli}}{H{\"o}lldobler and {Kencana Ramli}}{2009a}]{hk:2009a}
{\sc H{\"o}lldobler, S.} {\sc and} {\sc {Kencana Ramli}, C.~D.} 2009a.
\newblock Logic programs under three-valued {{\L}}ukasiewicz semantics.
\newblock In {\em Logic Programming, 25th International Conference,
  {ICLP~2009}}, {P.~M. Hill} {and} {D.~S. Warren}, Eds. Lecture Notes in
  Computer Science, vol. 5649. Springer, Heidelberg, 464--478.

\bibitem[\protect\citeauthoryear{H{\"o}lldobler and {Kencana
  Ramli}}{H{\"o}lldobler and {Kencana Ramli}}{2009b}]{hk:2009b}
{\sc H{\"o}lldobler, S.} {\sc and} {\sc {Kencana Ramli}, C.~D.} 2009b.
\newblock Logics and networks for human reasoning.
\newblock In {\em International Conference on Artificial Neural Networks,
  {ICANN~2009}, Part~II}, {C.~Alippi}, {M.~M. Polycarpou}, {C.~G. Panayiotou},
  {and} {G.~Ellinas}, Eds. Lecture Notes in Computer Science, vol. 5769.
  Springer, Heidelberg, 85--94.

\bibitem[\protect\citeauthoryear{H{\"o}lldobler, Philipp, and
  Wernhard}{H{\"o}lldobler et~al\mbox{.}}{2011}]{HPW:2011:AMHR}
{\sc H{\"o}lldobler, S.}, {\sc Philipp, T.}, {\sc and} {\sc Wernhard, C.} 2011.
\newblock An abductive model for human reasoning.
\newblock In {\em Logical Formalizations of Commonsense Reasoning, Papers from
  the AAAI 2011 Spring Symposium}. AAAI Spring Symposium Series Technical
  Reports. AAAI Press, Cambridge, MA, 135--138.

\bibitem[\protect\citeauthoryear{Kakas, Kowalski, and Toni}{Kakas
  et~al\mbox{.}}{1993}]{kakas:etal:93}
{\sc Kakas, A.~C.}, {\sc Kowalski, R.~A.}, {\sc and} {\sc Toni, F.} 1993.
\newblock Abductive logic programming.
\newblock {\em Journal of Logic and Computation\/}~{\em 2,\/}~6, 719--770.

\bibitem[\protect\citeauthoryear{Kowalski}{Kowalski}{2011}]{kowalski:2011}
{\sc Kowalski, R.} 2011.
\newblock {\em Computational Logic and Human Thinking: How to be Artificially
  Intelligent}.
\newblock Cambridge University Press, Cambridge.

\bibitem[\protect\citeauthoryear{{\L}ukasiewicz}{{\L}ukasiewicz}{1920}]{lukasiewicz:20}
{\sc {\L}ukasiewicz, J.} 1920.
\newblock O logice tr\'ojwarto\'sciowej.
\newblock {\em Ruch Filozoficzny\/}~{\em 5}, 169--171.
\newblock English translation: On three-valued logic. In: {\L}ukasiewicz J. and
  Borkowski L. (ed.). (1990). \textit{Selected Works}, Amsterdam: North
  Holland, pp. 87--88.

\bibitem[\protect\citeauthoryear{Pereira, Apar\'{i}cio, and Alferes}{Pereira
  et~al\mbox{.}}{1991}]{Pereira:wfs:iconstraints:91}
{\sc Pereira, L.~M.}, {\sc Apar\'{i}cio, J.~N.}, {\sc and} {\sc Alferes, J.~J.}
  1991.
\newblock Hypothetical reasoning with well founded semantics.
\newblock In {\em Scandinavian Conference on Artificial Intelligence: Proc.\ of
  the SCAI'91}, {B.~Mayoh}, Ed. IOS Press, Amsterdam, 289--300.

\bibitem[\protect\citeauthoryear{Pereira and Pinto}{Pereira and
  Pinto}{2011}]{pereira:pinto:2011}
{\sc Pereira, L.~M.} {\sc and} {\sc Pinto, A.~M.} 2011.
\newblock Inspecting side-effects of abduction in logic programs.
\newblock In {\em Logic Programming, Knowledge Representation, and Nonmonotonic
  Reasoning: Essays in honour of Michael Gelfond}, {M.~Balduccini} {and} {T.~C.
  Son}, Eds. LNAI, vol. 6565. Springer, 148--163.

\bibitem[\protect\citeauthoryear{Saptawijaya and Pereira}{Saptawijaya and
  Pereira}{2013}]{saptawijaya:pereira:2013}
{\sc Saptawijaya, A.} {\sc and} {\sc Pereira, L.~M.} 2013.
\newblock Tabled abduction in logic programs.
\newblock {\em Theory and Practice of Logic Programming\/}~{\em
  13,\/}~4-5-Online-Supplement.

\bibitem[\protect\citeauthoryear{Stenning and van Lambalgen}{Stenning and van
  Lambalgen}{2005}]{stenning:vanlambalgen:2005}
{\sc Stenning, K.} {\sc and} {\sc van Lambalgen, M.} 2005.
\newblock Semantic interpretation as computation in nonmonotonic logic: The
  real meaning of the suppression task.
\newblock {\em Cognitive Science\/}~{\em 6,\/}~29, 916--960.

\bibitem[\protect\citeauthoryear{Stenning and van Lambalgen}{Stenning and van
  Lambalgen}{2008}]{stenning:vanlambalgen:2008}
{\sc Stenning, K.} {\sc and} {\sc van Lambalgen, M.} 2008.
\newblock {\em Human Reasoning and Cognitive Science}.
\newblock A Bradford Book. MIT Press, Cambridge, MA.

\bibitem[\protect\citeauthoryear{Torasso, Console, Portinale, and
  Dupr{\'e}}{Torasso et~al\mbox{.}}{1991}]{torasso:console:luigi:1995}
{\sc Torasso, P.}, {\sc Console, L.}, {\sc Portinale, L.}, {\sc and} {\sc
  Dupr{\'e}, D.~T.} 1991.
\newblock On the relationship between abduction and deduction.
\newblock {\em Journal of Logic and Computation\/}~{\em 1,\/}~5, 661--690.

\bibitem[\protect\citeauthoryear{Van~Gelder, Ross, and Schlipf}{Van~Gelder
  et~al\mbox{.}}{1991}]{gelder:ross:schlipf:1991}
{\sc Van~Gelder, A.}, {\sc Ross, K.~A.}, {\sc and} {\sc Schlipf, J.~S.} 1991.
\newblock The well-founded semantics for general logic programs.
\newblock {\em Journal of the ACM\/}~{\em 38,\/}~3, 619--649.

\bibitem[\protect\citeauthoryear{Wason}{Wason}{1968}]{wason:68}
{\sc Wason, P.} 1968.
\newblock Reasoning about a rule.
\newblock {\em Quarterly Journal of Experimental Psychology\/}~{\em 20,\/}~3,
  273--281.

\end{thebibliography}

\label{lastpage}
\end{document}